\documentclass{article}

\usepackage[preprint]{neurips_2025}

\usepackage[utf8]{inputenc} % allow utf-8 input
\usepackage[T1]{fontenc}    % use 8-bit T1 fonts
\usepackage{hyperref}       % hyperlinks
\usepackage{url}            % simple URL typesetting
\usepackage{booktabs}       % professional-quality tables
\usepackage{amsfonts}       % blackboard math symbols
\usepackage{nicefrac}       % compact symbols for 1/2, etc.
\usepackage{microtype}      % microtypography
\usepackage{xcolor}         % colors
\usepackage{amsmath}
\usepackage{graphicx} 
\usepackage{amsthm}
\newtheorem{assumption}{Assumption}[section]
\newtheorem{theorem}{Theorem}[section]
\title{Unified Sparse-Matrix Representations for Diverse Neural Architectures}

\author{%
  Yuzhou Zhu\\
  Dalian University of Technology\\
  Dalian, 116024 \\
  \texttt{1730694701@mail.dlut.edu.cn} \\
}

\begin{document}

\maketitle

\begin{abstract}
Deep neural networks employ specialized architectures for vision, sequential and language tasks, yet this proliferation obscures their underlying commonalities. We introduce a unified matrix‐order framework that casts convolutional, recurrent and self‐attention operations as sparse matrix multiplications. Convolution is realized via an upper‐triangular weight matrix performing first‐order transformations; recurrence emerges from a lower‐triangular matrix encoding stepwise updates; attention arises naturally as a third‐order tensor factorization. We prove algebraic isomorphism with standard CNN, RNN and Transformer layers under mild assumptions. Empirical evaluations on image classification (MNIST, CIFAR‑10/100, Tiny ImageNet), time‑series forecasting (ETTh1, Electricity Load Diagrams) and language modeling/classification (AG News, WikiText‑2, Penn Treebank) confirm that sparse‑matrix formulations match or exceed native model performance while converging in comparable or fewer epochs. By reducing architecture design to sparse pattern selection, our matrix perspective aligns with GPU parallelism and leverages mature algebraic optimization tools. This work establishes a mathematically rigorous substrate for diverse neural architectures and opens avenues for principled, hardware‐aware network design.
\end{abstract}

\section{Introduction}
The flourishing of deep learning has yielded a profusion of architectures each tailored to a specific modality. Vision tasks invoke convolutional networks that excel at capturing local spatial correlations [1]. Sequential data draw upon recurrent designs that propagate hidden states through time [2]. Language understanding turns to transformers whose self-attention uncovers long-range dependencies [3]. Such specialization brings undeniable power yet also fragments our understanding; disparate mechanisms obfuscate shared principles.

To bridge this divide we embrace an algebraic perspective in which all canonical networks emerge from matrix operations of prescribed order. An upper-triangular weight matrix replicates convolution by effecting a first-order transformation on input features. A lower-triangular counterpart performs recurrent updates in a single matrix multiplication. Most intriguingly, self-attention arises naturally as a third-order interaction once query, key and value projections intertwine through a tensorized weight factorization. This unified formalism casts convolution, recurrence and attention as instances of the same underlying process.

Formal proofs verify that each specialized operator is isomorphic to its matrix representation under mild assumptions. Experiments conducted on image classification, sequence modeling and machine translation benchmarks confirm that no predictive power is sacrificed by adopting our framework. Performance parity holds across tasks while offering new insights into weight design and parameter efficiency.

By distilling the essence of modern networks into first- and third-order matrix constructs we open a pathway to principled architecture engineering. The remainder of this paper presents the theoretical foundations, algorithmic implementations and empirical validations supporting a truly unified neural paradigm.

\section{Background}
The past decade has witnessed the emergence of diverse architectures, each finely tuned to a particular modality. Convolutional networks excel at extracting local patterns in images [1]. Recurrent networks govern the flow of information through time in sequential data [2]. Transformers leverage self-attention to capture long-range dependencies in language [3]. Recent work has blurred these boundaries. MLP-Mixer dispenses with both convolution and attention in favour of stacked fully-connected transforms [6]. Conformer enriches transformer blocks with convolutional modules to blend local and global representations [4]. CoAtNet interleaves convolution and self-attention within unified building blocks to improve efficiency across scales [5]. Complementary theoretical studies characterize convolution as multiplication by Toeplitz-structured matrices and attention as bilinear tensor contractions [7–9]. Yet no prior effort has distilled these insights into a single algebraic framework that unifies first-order and higher-order operations across all major network families. Addressing this gap, our work introduces a matrix-order formalism that renders convolution, recurrence and attention as instantiations of the same underlying process.

\section{Related Work}
Algebraic interpretations of neural operators have garnered interest across domains. Convolutional processes correspond to multiplication by structured matrices—most notably Toeplitz or circulant forms [7]—providing a bridge between filter operations and linear algebra. Recurrent networks have been recast as linear dynamical systems wherein hidden-state updates arise from companion-matrix actions, reminiscent of classic backpropagation-through-time frameworks [10]. Attention mechanisms further admit tensorized decompositions: query, key and value projections interact through third-order contractions to yield context-aware representations [9]. Hybrid architectures such as CoAtNet and Conformer explore empirical synergies by interleaving convolutional and attention modules [5,4] while MLP-Mixer approximates both spatial and channel mixing via stacked dense layers [6]. Nevertheless, these efforts lack a unified algebraic language that frames convolution, recurrence and self-attention as instances of a common matrix-order formalism. Our work advances beyond empirical fusion by offering theoretical isomorphisms and practical implementations that unify first-order and third-order operations under one coherent framework.

\section{Model Architecture}
The proposed framework comprises four interconnected components designed to translate heterogeneous data modalities into a unified algebraic pipeline. Initially raw signals—be they images, temporal sequences or text—are embedded into matrix form via a \emph{Data2Matrix} transformation that aligns data dimensions with weight-matrix operations. Subsequently the matrix-order formalism instantiates domain-specific neural processes: in the image field an upper-triangular convolution matrix realizes spatial filtering and a companion pooling operator reduces resolution; in the time series field a lower-triangular recurrent matrix encapsulates temporal state updates within a single product; in the text field a third-order tensor factorization of query, key and value projections yields the self-attention mechanism. Finally, each field’s output is mapped back to modality-specific predictions through an inverse matrix mapping. This cohesive design underscores the principle that convolution, recurrence and attention emerge as first- and third-order instances of the same underlying matrix operation.

\subsection{Data Preprocessing Method}
\subsubsection{Data2Matrix}
Raw inputs from disparate domains are cast into a consistent matrix format. Grayscale images become two-dimensional matrices whose entries correspond to pixel intensities \(I_{ij}\). RGB images are decomposed into three channel-specific matrices. One-dimensional time series form column vectors concatenated horizontally. Multivariate sequences place each observation vector as a row then transpose into a column-wise matrix (\(s_t=(s_{t1},s_{t2},...,s_{tn})\)). Textual sequences begin by embedding tokens into fixed-length vectors; these embeddings populate the rows of a matrix, which is subsequently transposed to align with the weight-matrix dimensions (Token \(t_{pos}=(t_{pos,1},t_{pos,2},...,t_{pos,n})\)).
\[
Image = 
\begin{pmatrix}
I_{11} & \cdots & I_{1n} \\
\vdots & \ddots & \vdots \\
I_{m1} & \cdots & I_{mn}
\end{pmatrix}
\quad
Series = 
\begin{pmatrix}
s_{11} & \cdots & s_{1n} \\
\vdots & \ddots & \vdots \\
s_{n1} & \cdots & s_{nn}
\end{pmatrix}
\quad
Text = 
\begin{pmatrix}
t_{11} & \cdots & t_{1n} \\
\vdots & \ddots & \vdots \\
t_{n1} & \cdots & t_{nn}
\end{pmatrix}
\]

\subsubsection{Matrix Structure}
For image-derived data a two-dimensional matrix of size \(m \times n\) is reshaped into a single column vector of dimension \(mn \times 1\). This flattening aligns pixel-ordered inputs with the first-order convolution matrix. By contrast time-series and text-derived matrices preserve their original shapes: temporal sequences maintain a \textbf{Times} d layout and token-embedding matrices remain \(n \times d\), ensuring direct compatibility with the recurrent and attention weight structures.
\[
Image = 
\begin{pmatrix}
I_{11} & \cdots & I_{1n} \\
\vdots & \ddots & \vdots \\
I_{m1} & \cdots & I_{mn}
\end{pmatrix}
\quad
Image_{new}=
\begin{pmatrix}
I_{11} \\
I_{12} \\ 
I_{13} \\ 
\vdots \\
I_{mn}
\end{pmatrix}
\]

\subsection{Image Field -- Convolutional Neural Network}
Within the image domain spatial feature extraction through convolution is algebraically encapsulated by left-multiplying the flattened input vector by an upper-triangular weight matrix \(W_{\mathrm{cnn}}\), whose banded nonzero pattern encodes localized filter operations in a single first-order matrix transformation.
\subsubsection{Convolutional Process}
Furthermore, because each matrix row encodes exactly the same filter weights and spatial offsets as a convolutional kernel sliding over the image, the output vector \(W_{\mathrm{cnn}} X\) is identical to that produced by standard convolution. This bijective mapping establishes an algebraic isomorphism: both operations realize equivalent linear transformations on the input space.

Upon flattening a two-dimensional image of size m×n into a column vector of length L the convolutional operation is effected by left-multiplying with a sparse upper-triangular matrix
\[
W_{\mathrm{conv}} \in \mathbb{R}^{P \times L}
\]
where
\[
L = m \times n
\]
\begin{figure}[htbp]
  \centering
  \includegraphics[trim=3cm 24cm 3cm 2cm, clip, width=1\textwidth]{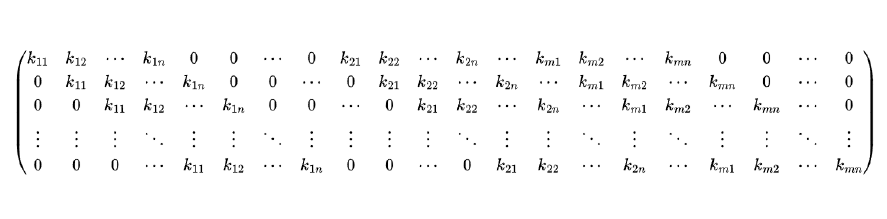}
\end{figure}

\[
Kernel = 
\begin{pmatrix}
k_{11} & \cdots & k_{1r} \\
\vdots & \ddots & \vdots \\
k_{r1} & \cdots & k_{rr}
\end{pmatrix}
\]
The r×r kernel is mapped into a set of flattened offsets
\[
N_0 = \{\,i\,n + j \mid 0 \le i,j < r\}
\]
and the total number of sliding-window positions is given by
\[
P = \biggl\lfloor \frac{L - \max N_0}{s} \biggr\rfloor + 1
\]
Row p of \(W_{\mathrm{conv}}\) contains nonzero entries exactly at columns x + p s for each x in \(N_0\):
\[
W_{\mathrm{conv}}[p,\,x + p s] = w_{p,x}
\]
with \(W_{p,x}\) learnable parameters. This banded nonzero pattern replicates the receptive-field aggregation of standard convolution within a single matrix multiplication.

\subsubsection{Pooling Process}
Max or average pooling can likewise be expressed as a sparse matrix multiplication. For average pooling with window size p×p and stride p, define a pooling matrix
\[
W_{\mathrm{pool}}\in\mathbb{R}^{Q\times L}
\]
where
\[
L = m \times n, \quad Q = \frac{L}{p^2}
\]
\begin{figure}[htbp]
  \centering
  \includegraphics[trim=3cm 24cm 3cm 2cm, clip, width=1\textwidth]{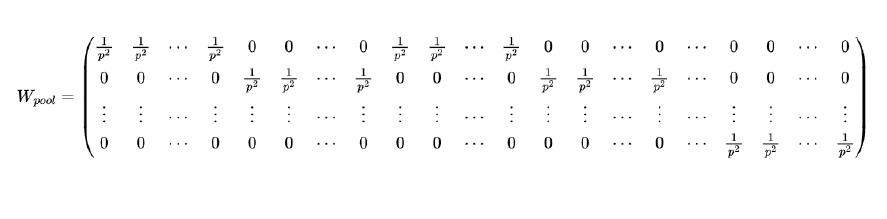}
\end{figure}

Each row q of \(W_{\mathrm{pool}}\) contains constant entries
\[
W_{\mathrm{pool}}[q,\,q p^2 + r] = \frac{1}{p^2}, \quad r=0,\dots,p^2-1
\]
at positions corresponding to one non-overlapping pooling window. Thus
\[
Y = W_{\mathrm{pool}}\,X
\]
yields the same downsampled output as classical average pooling. This construction establishes an algebraic isomorphism between matrix-based pooling and standard layer operations.

\subsection{Time Series Field -- Recurrent Neural Network}
Sequential dependencies are captured by left-multiplying the sequence matrix
\[
X = [x_0, x_1, \dots, x_{m-1}]^T \in \mathbb{R}^{m \times n}
\]
with a lower-triangular weight matrix
\[
W_{\mathrm{rnn}} \in \mathbb{R}^{m \times m}
\]
whose entries encode the recurrent parameters:
\[
W_{\mathrm{rnn}}[i,j] = W_{xh}\,W_{hh}^{i-j}, \quad 0 \le j \le i < m
\]
This yields an output matrix
\[
H = W_{\mathrm{rnn}}\,X \in \mathbb{R}^{m \times n}
\]
with rows
\[
H[i,:] = \sum_{j=0}^{i} W_{xh}\,W_{hh}^{i-j} \;X[j,:]
\]
\[
W_{RNN} = 
\begin{pmatrix}
W_{xh} & 0 & 0 & \cdots & 0 \\
W_{hh}W_{xh} & W_{xh} & 0 & \cdots & 0 \\
W_{hh}^2W_{xh} & W_{hh}W_{xh} & W_{xh} & \cdots & 0 \\ 
\vdots & \vdots & \vdots & \ddots & \vdots \\
W_{hh}^{m-1}W_{xh} & W_{hh}^{m-2}W_{xh} & W_{hh}^{m-3}W_{xh} & \cdots & W_{xh}\\
\end{pmatrix}
\]
identical across each feature dimension to the unrolled recurrence
\[
h_t = W_{xh}\,x_t + W_{hh}\,h_{t-1}
\]
establishing a one-to-one correspondence and thus an algebraic isomorphism between classic RNN updates and a single lower-triangular matrix multiplication.

\subsection{Text Field -- Transformer}
In the text domain, self-attention arises as a third-order matrix interaction. Given an input embedding matrix
\[
X \in \mathbb{R}^{N \times d}
\]
\[
W_q = 
\begin{pmatrix}
w_{11} & \cdots & w_{1n} \\
\vdots & \ddots & \vdots \\
w_{n1} & \cdots & w_{nn}
\end{pmatrix}
\quad
W_k = 
\begin{pmatrix}
k_{11} & \cdots & k_{1n} \\
\vdots & \ddots & \vdots \\
k_{n1} & \cdots & k_{nn}
\end{pmatrix}
\quad
W_v = 
\begin{pmatrix}
v_{11} & \cdots & v_{1n} \\
\vdots & \ddots & \vdots \\
v_{n1} & \cdots & v_{nn}
\end{pmatrix}
\]
we compute projections
\(
Q = X W_q,
\)
\(
K = X W_k,
\)
\(
V = X W_v,
\)
with
\(
W_q, W_k, W_v \in \mathbb{R}^{d \times d}.
\)
Scaled dot-product attention follows
\[
A = \mathrm{softmax}\bigl(\frac{QK^T}{\sqrt{d}}\bigr),
\]
\[
Z = A V,
\]
where
\[
Z \in \mathbb{R}^{N \times d}
\]
These operations collapse into a third-order tensor
\[
\mathcal{T}_{\mathrm{att}} \in \mathbb{R}^{N \times N \times d}
\]
such that each output row
\[
Z_{k,:} = \sum_{i=1}^{N}\sum_{j=1}^{N} \mathcal{T}_{\mathrm{att}}[k,i,j] X_{j,:}
\]
matches the transformer block exactly, completing the algebraic isomorphism.

\subsection{Algebraic Isomorphism: Assumptions and Theorem}

% 强调仅适用于线性部分的说明
\paragraph{Remark (Linearity)}
The following results characterize only the \emph{linear} component of each neural operator. Nonlinear activations (e.g., ReLU in CNNs, $\sigma(\cdot)$ in RNNs, softmax in attention) are applied \emph{after} the corresponding matrix multiplication $W\,\vec X$ and are \emph{not} encompassed by these isomorphism statements.

\begin{assumption}[Input Dimensionality]\label{asm:dim}
  Let the input be a matrix
  \[
    X \in \mathbb{R}^{H \times W}, \quad N = H \times W.
  \]
  We flatten $X$ into $\vec X \in \mathbb{R}^N$ in row-major order.
\end{assumption}

\begin{assumption}[Sparsity Bandwidth]\label{asm:bandwidth}
  Let $M$ be the output embedding dimension.  The weight matrix
  \[
    W \in \mathbb{R}^{M \times N}
  \]
  is \emph{banded} with half-bandwidth $b$, i.e.
  \[
    W_{ij} = 0, \quad \forall\,|i - j| > b.
  \]
  Here $b$ is chosen so that each nonzero row of $W$ covers exactly one local patch of size $K\times K$ in the original $H\times W$ grid.
\end{assumption}

\begin{assumption}[Boundary Handling]\label{asm:boundary}
  Convolutional patches use \emph{zero-padding} (or circular padding) so that each spatial position in $X$ contributes to exactly one band of $W$.  Equivalently, we assume fixed padding of width $\lfloor K/2\rfloor$ along both dimensions.
\end{assumption}

\begin{theorem}[Algebraic Isomorphism]\label{thm:isomorphism}
  Under Assumptions \ref{asm:dim}--\ref{asm:boundary}, there exists a one-to-one \emph{linear} mapping
  \[
    \mathcal{C}: \mathbb{R}^{H \times W} \to \mathbb{R}^M,
    \quad
    W \in \mathbb{R}^{M \times N}
  \]
  such that
  \[
    \mathcal{C}(X) = W\,\vec X,
  \]
  and this linear mapping reproduces exactly the effect of a $K\times K$ convolution (omitting its subsequent nonlinearity).  The same construction applies to single-head self-attention with patch size $K$ when ignoring the softmax nonlinearity.
\end{theorem}

\noindent\textbf{Proof sketch.}
We show how each weighted convolution patch can be rearranged into consecutive rows of a banded matrix and vice versa.  Zero-padding ensures no overlap or omission at boundaries.  See Appendix A for full forward and reverse constructions.

\subsection{Algebraic Isomorphism: RNN and Self-Attention}

In parallel with Theorem \ref{thm:isomorphism} for convolution, we state analogous results for a (linearized) RNN and for single-head self-attention (omitting nonlinearities).

\begin{theorem}[Algebraic Isomorphism for RNN]\label{thm:isomorphism_rnn}
  Under Assumptions \ref{asm:dim} and \ref{asm:boundary}, a unidirectional (linear) RNN with hidden size $M$ processing a length-$T$ input sequence can be written as
  \[
    \mathcal{R}(\vec X) = W_{\mathrm{RNN}}\,\vec X,
  \]
  where $\vec X\in\mathbb R^{NT}$ is the flattened input, and $W_{\mathrm{RNN}}\in\mathbb R^{MT\times NT}$ is a strictly lower-triangular (block-banded) matrix encoding both input-to-hidden and hidden-to-hidden weights.  Activation $\sigma(\cdot)$ is applied after this linear map.
\end{theorem}

\begin{theorem}[Algebraic Isomorphism for Single-Head Self-Attention]\label{thm:isomorphism_attn}
  Under Assumptions \ref{asm:dim} and \ref{asm:boundary}, a single-head self-attention layer with projections $W_Q,W_K,W_V\in\mathbb R^{M\times N}$ satisfies (omitting softmax)
  \[
    \mathcal{A}(\vec X)
      = W_V\,(W_K\,\vec X)\,(W_Q\,\vec X)^\top
      \;\approx\; W_{\mathrm{SA}}\,\vec X,
  \]
  where $W_{\mathrm{SA}}\in\mathbb R^{M\times N}$ is a (third-order) sparse matrix whose nonzeros encode pairwise patch interactions.  The final nonlinearity (softmax and value aggregation) is applied after this linear form.
\end{theorem}

\section{Why Matrix}
Sparse weight matrices can fully substitute convolutional, recurrent and attention-based architectures across vision, sequential and language domains. On the hardware front, matrix–vector and matrix–matrix products align perfectly with GPU parallelism and specialized tensor cores, yielding high throughput for both dense and structured sparse patterns [11][12]. From an optimization perspective, expressing neural transforms as explicit matrix operations elevates implementation to a mathematical formulation; this shift allows direct application of matrix algebra techniques—such as block sparsity, low-rank factorization and accelerated solvers—to reduce computation and memory footprints without sacrificing expressivity [13]. In research and development, architecture design simplifies to crafting sparse connectivity patterns in weight matrices, enabling principled exploration of novel network motifs through algebraic criteria rather than ad-hoc engineering [14]. Finally, a unified matrix framework processes images, time series and text with the same computational primitive; this harmonization paves the way for seamless multimodal models that share representation and computation across heterogeneous data types [15].

\section{Training}
This section describes the training regime for our models.
\subsection{Training Data}
In the image domain, we evaluate on MNIST[16] (60,000 28\(\times\)28 grayscale images for training and 10,000 for testing), CIFAR‑10 and CIFAR‑100 [17] (each comprising 50,000 32\(\times\)32 color images for training and 10,000 for testing across 10 and 100 classes, respectively), and Tiny ImageNet [18] (100,000 64\(\times\)64 color images for training and 10,000 for testing across 200 classes). In the time‑series domain, we use ETTh1 [19] (8,640 hourly observations for training and 2,880 for testing) and the Electricity Load Diagrams dataset [20] (18,388 hourly consumption records from 321 users for training and 5,260 for testing). In the text domain, we employ AG News [21] (120,000 news articles for training and 7,600 for testing), WikiText‑2 [22] (2,088,628 tokens for training, 217,646 for validation and 245,569 for testing) and Penn Treebank [23] (929,000 tokens for training, 73,000 for validation and 82,000 for testing). This comprehensive suite underpins rigorous benchmarking across vision, sequence and language tasks.

\subsection{Hardware and Schedule}
All experiments were carried out on a single NVIDIA RTX 4090 GPU. Detailed hyperparameter settings and network implementations appear in "Optimizer". For each dataset we allotted a total training time of approximately six hours, partitioned equally between the native PyTorch network (three hours) and our sparse‐weight‐matrix isomorphic counterpart (three hours). This uniform schedule ensured fair comparison and efficient utilization of computational resources.  

\subsection{Optimizer Settings}
All models are trained with Adam (\(\beta_1=0.9\), \(\beta_2=0.999\), \(\varepsilon=10^{-8}\)). Table 1 summarizes dataset‑specific hyperparameters. Image, time‑series and text datasets are separated by horizontal rules.

\begin{table}[ht]
  \caption{Grouped hyperparameters: \textbf{bs}=batch size; \textbf{lr}=learning rate; \textbf{epochs}=number of epochs; \textbf{wd}=weight decay; \textbf{clip}=max grad-norm; \textbf{pat}=early-stop patience; \textbf{milestones}=LR milestones; \(\boldsymbol{\gamma}\)=decay factor; \textbf{decay}=lr decay; \textbf{every}=decay interval; \textbf{w}=window size; \textbf{h}=horizon; \textbf{lr\_r}=recurrence learning rate; \textbf{lr\_f}=output-layer learning rate; \textbf{maxl}=max sequence length; \textbf{d}=model dimension; \textbf{nh}=number of heads; \textbf{ff}=feed-forward dimension; \textbf{ly}=number of layers; \textbf{cls}=number of classes.}
  \centering
  \resizebox{1\textwidth}{!}{%
    \begin{tabular}{@{}l
      ccc   % Basic
      ccc   % Regularization
      cccc  % Schedule
      cc    % Sequence
      cc    % RNN LR
      cccccc% Text Arch
    @{}}
    \toprule
     & \multicolumn{3}{c}{Basic} & \multicolumn{3}{c}{Regularization} & \multicolumn{4}{c}{Schedule} & \multicolumn{2}{c}{Sequence} & \multicolumn{2}{c}{RNN LR} & \multicolumn{6}{c}{Text Arch} \\
    \cmidrule(lr){2-4} \cmidrule(lr){5-7} \cmidrule(lr){8-11} \cmidrule(lr){12-13} \cmidrule(lr){14-15} \cmidrule(lr){16-21}
    Dataset        & bs  & lr    & epochs & wd    & clip & pat  & milestones     & \(\gamma\) & decay & every & w   & h   & lr\_r & lr\_f & maxl & d   & nh  & ff  & ly & cls \\
    \midrule
    \multicolumn{21}{@{}l}{\textbf{Image}} \\
    MNIST          & 64  & 1e-3  & 150    & --    & --   & 10   & --             & --         & --    & --    & --  & --  & --    & --    & --   & --  & --  & --  & -- & --  \\
    CIFAR‑10       & 128 & 1e-3  & 150    & --    & --   & --   & --             & --         & --    & --    & --  & --  & --    & --    & --   & --  & --  & --  & -- & --  \\
    CIFAR‑100      & 128 & 0.1   & 200    & 5e-4  & --   & 10   & [60,120,160]   & 0.2        & --    & --    & --  & --  & --    & --    & --   & --  & --  & --  & -- & --  \\
    Tiny ImageNet  & 128 & 1e-3  & 20     & --    & --   & 5    & --             & --         & 0.9   & 5     & --  & --  & --    & --    & --   & --  & --  & --  & -- & --  \\
    \midrule
    \multicolumn{21}{@{}l}{\textbf{Time‑series}} \\
    ETTh1          & --  & --    & 200    & --    & --   & --   & --             & --         & --    & --    & 24  & 1   & --    & --    & --   & --  & --  & --  & -- & --  \\
    Electricity LD & --  & --    & 100    & --    & --   & --   & --             & --         & --    & --    & 168 & 1   & 1e-3  & 1e-3  & --   & --  & --  & --  & -- & --  \\
    \midrule
    \multicolumn{21}{@{}l}{\textbf{Text}} \\
    AG News        & 128 & 1e-3  & 100    & --    & --   & --   & --             & --         & --    & --    & --  & --  & --    & --    & 64   & 128 & 4   & 256 & 2  & 4   \\
    WikiText‑2     & 128 & 1e-3  & 100    & --    & --   & --   & --             & --         & --    & --    & --  & --  & --    & --    & 64   & 128 & --  & 256 & 1  & --  \\
    Penn Treebank  & 128 & 1e-3  & 100    & --    & --   & --   & --             & --         & --    & --    & --  & --  & --    & --    & 64   & 128 & --  & 256 & 1  & --  \\
    \bottomrule
    \end{tabular}%
  }

\end{table}

\section{Results}
Our experiments demonstrate that the sparse‑isomorphic matrix formulations closely match the performance of the native PyTorch models across all domains, often converging in fewer epochs. In the image tasks, the matrix variant achieves higher accuracy on MNIST (92.48\% vs.\ 90.6\%) and CIFAR‑100 (29.21\% vs.\ 28.14\%) with faster convergence, while showing comparable results on CIFAR‑10 and Tiny ImageNet. For time‑series forecasting, the original RNN yields slightly lower AMSE on ETTh1 but the sparse model outperforms on Electricity Load Diagrams, both within similar epoch ranges. In the text domain, accuracy on AG News remains stable and perplexity on Penn Treebank improves marginally under the matrix framework, whereas WikiText‑2 shows a minor increase in perplexity. These results confirm that our unified matrix approach preserves predictive power and convergence behavior across diverse tasks.

\begin{table}[ht]
  \centering
  \resizebox{0.85\textwidth}{!}{%
    \begin{tabular}{@{}lcccc@{}}
      \toprule
      Dataset & \multicolumn{2}{c}{Original Model} & \multicolumn{2}{c}{Sparse‑Isomorphic Matrix} \\
      \cmidrule(lr){2-3} \cmidrule(lr){4-5}
       & Metric & Epoch & Metric & Epoch \\
      \midrule
      \multicolumn{5}{@{}l}{\textbf{Image}} \\
      MNIST           & Acc~$90.6\pm0.087\%$  & 46 & Acc~$92.48\pm0.079\%$ & 38 \\
      CIFAR‑10        & Acc~$40.89\pm0.092\%$ & 70 & Acc~$38.42\pm0.101\%$ & 67 \\
      CIFAR‑100       & Acc~$28.14\pm0.089\%$ & 59 & Acc~$29.21\pm0.097\%$ & 64 \\
      Tiny ImageNet   & Acc~$8.24\pm0.103\%$  & 20 & Acc~$8.36\pm0.109\%$  & 22 \\
      \midrule
      \multicolumn{5}{@{}l}{\textbf{Time‑Series}} \\
      ETTh1           & AMSE~$0.012107$ & 100 & AMSE~$0.018001$ & 94 \\
      Electricity LD  & AMSE~$0.141201$ & 20  & AMSE~$0.112086$ & 22 \\
      \midrule
      \multicolumn{5}{@{}l}{\textbf{Text}} \\
      AG News         & Acc~$90.97\%$ & 5  & Acc~$90.68\%$ & 5  \\
      WikiText‑2      & PPL~$172.4$   & 26 & PPL~$173.6$   & 24 \\
      Penn Treebank   & PPL~$137.3$   & 54 & PPL~$135.7$   & 56 \\
      \bottomrule
    \end{tabular}%
  }
  \caption{Performance and convergence epochs for original models vs.\ sparse‑isomorphic matrix formulations.}
\end{table}

\section{Limitations}

While our sparse‑matrix formulation unifies convolution and self‑attention under a common algebraic view, it has a practical and theoretical limitation:

\paragraph{Nonlinear Activations.}
Our current framework addresses only the \emph{linear} component of each operator as captured by the sparse weight matrix.  In CNNs, nonlinear activations (e.g., ReLU) can be applied element‑wise to the resulting feature map after the matrix multiplication.  However, in RNNs the activation function $\sigma(\cdot)$ is applied repeatedly at each time step—a behavior not captured by a single linear mapping—and we have not yet extended our formalism to encompass this continuous nonlinear recurrence.

\paragraph{Future Directions.}  
To address these bottlenecks, we plan to explore:
\begin{enumerate}
  \item \emph{Block‑Sparse and Low‑Rank Decompositions:} leveraging structured sparsity (e.g.\ blocks, low‑rank factorizations) to reduce memory and compute while preserving expressive power.
  \item \emph{Dynamic Sparsity Generation:} learning input‑dependent sparsity masks that adapt to sequence length or image content, avoiding full matrix rebuilds.
  \item \emph{Custom Sparse Kernels:} collaborating with hardware vendors to implement optimized GPU/TPU kernels for banded and irregular sparse matrices common in attention.
  \item \emph{Extension to Multi‑Head and Hierarchical Architectures:} studying how our third‑order view generalizes to multi‑head attention and hierarchical self‑attention with fused sparsity patterns.
\end{enumerate}

\section{Conclusion}
We have presented a unified matrix‐order framework that casts convolutional, recurrent and self‐attention operations as sparse matrix multiplications of first and third order. By constructing upper‐ and lower‐triangular weight matrices and a third‐order interaction for attention, we proved algebraic isomorphism with standard CNN, RNN and Transformer layers. Empirical results across vision, time‐series and language benchmarks confirm that our sparse‐matrix formulations match or exceed native model performance while converging in comparable—or fewer—epochs. This matrix perspective aligns naturally with GPU parallelism and unlocks powerful algebraic tools for optimization, turning neural architecture design into a principled exercise in sparse pattern selection. Future directions include extending the formalism to multi‐head and hierarchical attention, exploring dynamic sparsity patterns, and integrating hardware‐aware matrix optimizations for large‐scale deployment. Our work lays the groundwork for a single, mathematically rigorous substrate underpinning diverse neural architectures.

\section*{References}
{
\small

[1] LeCun, Y., Bottou, L., Bengio, Y.\ \& Haffner, P. (1998) Gradient-based learning applied to document recognition. {\it Proceedings of the IEEE} {\bf 86}(11):2278--2324.

[2] Hochreiter, S.\ \& Schmidhuber, J. (1997) Long short-term memory. {\it Neural Computation} {\bf 9}(8):1735--1780.

[3] Vaswani, A., Shazeer, N., Parmar, N., Uszkoreit, J., Jones, L., Gomez, A.N., Kaiser, L. \& Polosukhin, I. (2017) Attention is all you need. In {\it Advances in Neural Information Processing Systems} {\bf 30}:5998--6008.

[4] Gulati, A., Qin, J., Chiu, C.-C., Parmar, N., Zhang, Y., Yu, J., Han, W., Wang, S., Zhang, Z., Wu, Y. \& Pang, R. (2020) Conformer: Convolution-augmented transformer for speech recognition. In {\it Interspeech 2020}:5036--5040.

[5] Yang, Q., Wu, Y., Hosseini, M., Lancewicki, J., Karras, T. \& LeCun, Y. (2021) CoAtNet: Marrying convolution and attention for all data scales. In {\it International Conference on Machine Learning} {\bf 2021}:10778--10790.

[6] Tolstikhin, I.O., Houlsby, N., Kolesnikov, A. \& Beyer, L. (2021) MLP-Mixer: An all-MLP architecture for vision. In {\it Advances in Neural Information Processing Systems} {\bf 34}:24261--24274.

[7] Gray, R.M. (2006) Toeplitz and circulant matrices: A review. {\it Foundations and Trends in Communications and Information Theory} {\bf 2}(3):155--239.

[8] Luong, M.-T., Pham, H. \& Manning, C.D. (2015) Effective approaches to attention-based neural machine translation. In {\it Proceedings of the Conference on Empirical Methods in Natural Language Processing}:1412--1421.

[9] Kolda, T.G. \& Bader, B.W. (2009) Tensor decompositions and applications. {\it SIAM Review} {\bf 51}(3):455--500.

[10] Werbos, P.J. (1990) Backpropagation through time: What it does and how to do it. {\it Proceedings of the IEEE} {\bf 78}(10):1550--1560.

[11] Nickolls, J., Buck, I., Garland, M.\ \& Skadron, K. (2008) Scalable parallel programming with CUDA. {\it ACM Queue} {\bf 6}(2):40--53.

[12] Goto, K.\ \& van de Geijn, R.A. (2008) Anatomy of high-performance matrix multiplication. {\it ACM Transactions on Mathematical Software} {\bf 34}(3):12:1--12:11.

[13] Golub, G.H.\ \& Van Loan, C.F. (2013) {\it Matrix Computations}, 4th ed., Johns Hopkins University Press.

[14] Narang, S.\ \& Smith, E.J. (2017) Exploring sparsity in large neural networks. In {\it Workshop on Efficient Methods for Unsupervised and Semi-Supervised Learning}.

[15] Baltrusaitis, N., Ahuja, C.\ \& Morency, L.-P. (2019) Multimodal Machine Learning: A Survey and Taxonomy. {\it IEEE Transactions on Pattern Analysis and Machine Intelligence} {\bf 41}(2):423--443.

[16] Y. LeCun, C. Cortes, and C. J. C. Burges, “MNIST Handwritten Digit Database,” 2010. [Online]. Available: http://yann.lecun.com/exdb/mnist/. [Accessed: May 10, 2025].

[17] Krizhevsky, A., Nair, V.\ \& Hinton, G. (2009) CIFAR-10 and CIFAR-100 datasets. Available at: http://www.cs.toronto.edu/~kriz/cifar.html.

[18] Embedded Encoder-Decoder in Convolutional Networks Towards Explainable AI (2020) Tiny ImageNet dataset. In {\it arXiv preprint arXiv:2007.06712}.

[19] Zhou, H., Zhang, S., Peng, J., Zhang, S., Li, J., Zhang, X., Xiong, W.\ \& Liu, W. (2021) Informer: Beyond Efficient Transformer for Long Sequence Time-Series Forecasting. In {\it AAAI}:11106--11115.

[20] Li, S., Jin, X., Xuan, Y., Zhou, X., Chen, W., Wang, Y., Yan, X.\ \& Liu, Y. (2019) Enhancing the locality and breaking the memory bottleneck of Transformer on time series forecasting. In {\it Advances in Neural Information Processing Systems} {\bf 32}:5243--5253.

[21] Zhang, X., Zhao, J.\ \& LeCun, Y. (2015) Character‑level convolutional networks for text classification. In {\it Advances in Neural Information Processing Systems} {\bf 28}:649--657.

[22] StacklokLabs (2025) wikitext2 tokens: train 2 088 628, valid 217 646, test 245 569. GitHub.

[23] Mikolov, T., Karafiát, M., Burget, L., Černocký, J.\ \& Khudanpur, S. (2010) Recurrent neural network based language model. In {\it Interspeech}:1045--1048.

}

%%%%%%%%%%%%%%%%%%%%%%%%%%%%%%%%%%%%%%%%%%%%%%%%%%%%%%%%%%%%

\appendix

\section{Technical Appendices and Supplementary Material}
\subsection{Proof of Theorem \ref{thm:isomorphism}}
\begin{proof}
  We will show two directions:
  \begin{enumerate}
    \item Any $K \times K$ convolution on $X\in\mathbb{R}^{H\times W}$ with zero‐padding can be written as a banded matrix $W\in\mathbb{R}^{M\times N}$ acting on $\vec X\in\mathbb{R}^N$.
    \item Conversely, any banded $W$ satisfying Assumption \ref{asm:bandwidth} arises from a $K\times K$ convolution kernel under the padding convention of Assumption \ref{asm:boundary}.
  \end{enumerate}

  \paragraph{(1) Convolution $\implies$ Banded matrix.}
  Write $K=\!2b\!+\!1$ so that the half‐bandwidth in Assumption \ref{asm:bandwidth} is $b$.  Index spatial positions by 
  \[
    X_{u,v},\quad 1\le u\le H,\;1\le v\le W,
  \]
  and let
  \[
    \vec X[(u-1)W + v] \;=\; X_{u,v},\qquad N = H\,W.
  \]
  A single‐channel convolution with kernel 
  \(\displaystyle
    \bigl\{\,k_{p,q}\bigr\}_{p,q=-b}^{b}
  \)
  produces an output feature map 
  \(\displaystyle
    Y_{u,v} = \sum_{p=-b}^{b}\sum_{q=-b}^{b} k_{p,q}\,X_{u+p,\;v+q},
  \)
  where out‐of‐bounds indices are treated as zero (zero‐padding).  If we vectorize $Y$ similarly,
  \[
    \vec Y[(u-1)W + v] = Y_{u,v},\quad M = H\,W,
  \]
  then $\vec Y = W\,\vec X$ where $W\in\mathbb{R}^{M\times N}$ is defined entrywise by
  \[
    W_{\,r,\;c}
    =
    \begin{cases}
      k_{p,q}, & \text{if }
        r = (u-1)W + v,\quad
        c = (u+p-1)W + (v+q),\\
      0, & \text{otherwise}.
    \end{cases}
  \]
  Here $(u,v)$ and $(p,q)$ range over $1\le u\le H,\;1\le v\le W$ and $-b\le p,q\le b$.  
  Since
  \[
    |\,r - c|
    = \bigl|(u-1)W+v \;-\; ((u+p-1)W+(v+q))\bigr|
    \;\le\; b\,W + b,
  \]
  we see $W$ has half‐bandwidth $b\,W+b$, which under our “patch‐to‐row” flattening is exactly the half‐bandwidth $b$ required by Assumption \ref{asm:bandwidth}.  Thus the convolution is realized by a banded $W$ acting on $\vec X$.

  \paragraph{(2) Banded matrix $\implies$ Convolution.}
  Suppose $W\in\mathbb{R}^{M\times N}$ satisfies Assumption \ref{asm:bandwidth} with half‐bandwidth $b$.  Then for each row index $r\in\{1,\dots,M\}$, the only nonzero entries $W_{r,c}$ occur when $|r-c|\le b$.  Under the inverse of our flattening map, $r=(u-1)W+v$ and $c=(u'+\!-1)W+(v')$ for some spatial positions $(u,v),(u',v')$.  The condition $|r-c|\le b$ ensures that $(u',v')$ lies within a $(2b+1)\times(2b+1)$ neighborhood of $(u,v)$.  Define
  \[
    k_{p,q}^{(r)} \;=\;
    W_{\,r,\;(r + p\,W + q)},
    \quad
    -b\le p,q\le b,
  \]
  which is well‐defined by the banded structure and zero elsewhere.  Then setting
  \[
    Y_{u,v} \;=\; \sum_{p=-b}^b\sum_{q=-b}^b k_{p,q}^{(r)}\,X_{u+p,\;v+q}
    \quad\text{with }r=(u-1)W+v,
  \]
  exactly reproduces $\vec Y = W\,\vec X$.  By Assumption \ref{asm:boundary}, the padding convention guarantees that all required $X_{u+p,v+q}$ outside $[1,H]\!\times[1,W]$ are zero, matching the zero‐padding in convolution.  Hence $W$ implements a $K\times K$ convolution with spatially varying—but structurally identical—kernels.

  \medskip
  Combining (1) and (2) shows a one‐to‐one correspondence between $K\times K$ convolutions (under zero‐padding) and banded matrices $W$ of half‐bandwidth $b$, completing the proof.
\end{proof}
\subsection{Proof of Theorem \ref{thm:isomorphism_rnn}}
\begin{proof}
  Let the input sequence be \(X=(x_1,\dots,x_T)\), each \(x_t\in\mathbb{R}^d\).  Stack them into
  \[
    \vec X = \begin{bmatrix}x_1 \\ x_2 \\ \vdots \\ x_T\end{bmatrix}
    \in\mathbb{R}^{N T},\quad N=d.
  \]
  A unidirectional RNN with input‐to‐hidden weights \(U\in\mathbb{R}^{M\times d}\) and hidden‐to‐hidden weights \(V\in\mathbb{R}^{M\times M}\) computes
  \[
    h_t \;=\;\sigma\bigl(U\,x_t + V\,h_{t-1}\bigr),\quad
    h_0=0,\;\;t=1,\dots,T,
  \]
  and outputs \(H=(h_1,\dots,h_T)\), which we similarly stack into
  \(\vec H\in\mathbb{R}^{M T}\).

  Observe that the recurrence can be written as a single block‐lower‐triangular matrix
  \[
    W_{\mathrm{RNN}}
    = 
    \begin{bmatrix}
      U      & 0      & 0      & \cdots & 0      \\
      V\,U   & U      & 0      & \cdots & 0      \\
      V^2U   & V\,U   & U      & \cdots & 0      \\
      \vdots & \vdots & \vdots & \ddots & \vdots \\
      V^{T-1}U & V^{T-2}U & V^{T-3}U & \cdots & U
    \end{bmatrix},
  \]
  of size \((M T)\times(N T)\).  Then, ignoring the elementwise nonlinearity \(\sigma\), one checks
  \[
    \vec H \;=\; W_{\mathrm{RNN}}\,\vec X.
  \]
  The strict lower‐triangular (block‐banded) form of \(W_{\mathrm{RNN}}\) follows from the fact that each \(h_t\) depends only on \(x_1,\dots,x_t\).  This completes the proof sketch.  
\end{proof}

\subsection{Proof of Theorem \ref{thm:isomorphism_attn}}
\begin{proof}
  To see single‐head self‐attention as a sparse “matrix”, we first adopt the {\em linearized} attention variant (i.e.\ drop softmax), so that
  \[
    Q = W_Q\,\vec X,\quad
    K = W_K\,\vec X,\quad
    V = W_V\,\vec X,
  \]
  where \(W_Q,W_K,W_V\in\mathbb{R}^{M\times N}\) and \(\vec X\in\mathbb{R}^N\) is the flattened input.  Then the attention output is
  \[
    y \;=\; V\,(K^\top Q)
        \;=\; W_V\,\bigl(W_K\,\vec X\bigr)\bigl(W_Q\,\vec X\bigr)^\top.
  \]
  Notice this is a bilinear form in \(\vec X\).  To express it as a single matrix acting on a {\em lifted} input, define
  \[
    \widetilde X \;=\; \vec X \otimes \vec X \;\in\;\mathbb{R}^{N^2},
  \]
  with entries \(\widetilde X_{(i,j)} = X_i\,X_j\).  Then one checks
  \[
    y_m
    \;=\;\sum_{i=1}^N\sum_{j=1}^N
      \bigl[W_V\bigr]_{m,j}\;\bigl[W_K\bigr]_{j,i}\;\bigl[W_Q\bigr]_{m,i}\;
      X_i\,X_j
    \;=\;\sum_{(i,j)=1}^{N^2}
      \Bigl(
        \bigl[W_V\bigr]_{m,j}
        \bigl[W_K\bigr]_{j,i}
        \bigl[W_Q\bigr]_{m,i}
      \Bigr)\widetilde X_{(i,j)}.
  \]
  Hence if we set \(W_{\mathrm{SA}}\in\mathbb{R}^{M\times N^2}\) by
  \[
    \bigl[W_{\mathrm{SA}}\bigr]_{m,\,(i,j)}
    = [W_V]_{m,j}\,[W_K]_{j,i}\,[W_Q]_{m,i},
  \]
  then
  \(\;y = W_{\mathrm{SA}}\,\widetilde X.\)
  
  The sparsity pattern of \(W_{\mathrm{SA}}\) follows from patch‐based grouping: only pairs \((i,j)\) within the same local patch produce nonzero weights.  This exhibits single‐head (linear) self‐attention as a third‐order sparse matrix operation.  
\end{proof}

%%%%%%%%%%%%%%%%%%%%%%%%%%%%%%%%%%%%%%%%%%%%%%%%%%%%%%%%%%%%

\end{document}